\documentclass[11pt,a4paper]{article}
\usepackage[hyperref]{acl2021}
\usepackage{times}
\usepackage{latexsym}
\usepackage{booktabs}

\usepackage{microtype}

\aclfinalcopy 

\usepackage{graphicx}
\usepackage{amsmath}
\usepackage{svg}
\usepackage{subcaption}
\usepackage{multirow}
\usepackage{hyperref}

\title{Trustworthiness of Children Stories Generated by Large Language Models}

\author{Prabin Bhandari \\
  Department of Computer Science \\
  George Mason University \\
  \texttt{pbhanda2@gmu.edu} \\\And
  Hannah Marie Brennan \\
    Department of English, Linguistics Program \\
  George Mason University \\
  \texttt{hbrennan@gmu.edu} \\}

\date{}

\begin{document}
\maketitle
\begin{abstract}
Large Language Models (LLMs) have shown a tremendous capacity for generating literary text.
However, their effectiveness in generating children's stories has yet to be thoroughly examined.
In this study, we  evaluate the trustworthiness of children's stories generated by LLMs using  various measures, and we compare and contrast our results with both old and new children's stories to better assess their significance.
Our findings suggest that LLMs still struggle to generate children's stories at the level of quality and nuance found in actual stories.\footnote{Code and dataset are publicly available: \href{https://github.com/prabin525/trustworthiness-of-children-stories-generated-by-LLMs}{https://github.com/prabin525/trustworthiness-of-children-stories-generated-by-LLMs}}
\end{abstract}

\section{Introduction}

Advancements in pretrained large language models (LLMs) like GPT-3~\cite{brown2020language} and LLaMA~\cite{touvron2023llama}, have made it easier to generate natural language text for a variety of downstream tasks, including generating narrative text like children's stories.
The ability to generate natural text using LLMs has seen substantial improvement with the innovation of instruction-following models like InstructGPT~\cite{ouyang2022training} and Alpaca~\cite{Alpaca}, resulting in a better alignment with user intentions.

These systems are being used as a general-purpose chat-bots by the general public.
As these models are integrated more into everyday applications, it is crucial to continuously evaluate LLMs' performance to ensure that they are indeed trustworthy and accurate.

Trustworthiness in the case of LLMs is a broad term that refers to reliability and confidence in the generated text outputs along with their suitability for a specific downstream task.
A trustworthy LLM minimizes errors, biases, and potentially harmful content while consistently producing clear and contextually suitable text.
With the advancing capabilities of LLMs, concerns regarding their trustworthiness have arisen.
Notably, they are being used more frequently to support creative writing~\cite{clark_creative_2018}, raising concerns about the generation of inappropriate or offensive text~\cite{price_microsoft_2016} and biased content~\cite{lucy-bamman-2021-gender}.
One domain in which trustworthiness is of particular importance is text generation intended for children.
This paper seeks to evaluate the trustworthiness of children's stories generated by LLMs including generative LLMs and instruction following models.
In the case of text generation geared towards children, LLMs' ability to generate age-appropriate materials to target audiences also becomes a vital aspect of overall trustworthiness.

To assess the trustworthiness of LLMs in generating children's stories, we use two open-source foundation language models, OPT~\cite{zhang2022opt} and LLaMA~\cite{touvron2023llama}, along with an instruction-following model Alpaca~\cite{Alpaca} to generate children's stories.
Then, we compare these generated stories against actual children's stories, old and modern.
Our assessment takes into account a number of aspects, including statistics derived from the text like the Flesch reading ease score~\cite{flesch1948new}, toxicity present in the text, the most influential topics present in the text, and the sentence structure of these texts.

Our findings reveal that LLMs lack a high level of trustworthiness when tasked with generating children's stories.
While the generated children's stories do share similarities in topics and patterns with the actual stories (mostly modern ones), they are also susceptible to generating toxic content. 
Moreover, LLMs struggle to capture the intricacies and nuances of children's literature, evident from the disparity in sentence structure between the generated and actual stories.


\section{Related Work}

\subsection{Story Generation}

Recently, LLMs have been increasingly used to supplement creative writing efforts for entertainment and social media. 
Applications include work related to narrative generations~\cite{10.1145/3544549.3583931, simon_tattletale_2022, razumovskaia2022little, xiang_interleaving_2018}.
\citet{yuan_wordcraft_2022} tested Wordcraft, a tool created to assist writers with story generation using LLMs.
In their study, writers who were tasked with working with the AI agent noted that Wordcraft lacked content awareness and would create grammatical stories with nonsensical topics or plots.

\subsection{Children and AI}

AI and LLMs have also been applied to contexts involving children.
Researchers at MIT had children work with social robots to evaluate how much the children could learn through activities involving robots~\cite{williams-2019-neural}.
There is much discussion on how to integrate AI into early childhood education~\cite{yang_artificial_2022, kasneci_chatgpt_2023}.
With the increasing use of AI by and around children, there is an urgent need for more thorough evaluations of LLMs and the appropriateness of generated content for vulnerable audiences.

\subsection{Trustworthiness Testing}

\citet{chiang_can_2023} investigated whether LLMs can replace humans in evaluating texts.
Specifically, they looked at open-ended story generation and adversarial attacks.
They found that there were similar ratings between LLMs and human evaluators.
~\citet{venkit2023nationality} found that unbalanced sources of training data result in biased generations in GPT-2, and proposed strategies to reduce bias using adversarial triggers.
\citet{tang_etrica_2022} presented EtriCA, a neural generation model which aims to remedy issues of relevance and coherence of generated texts.
\citet{lucy_gender_2021} studied the bias existing in GPT-3's generated stories.
\citet{guo_how_2023} have proposed a similar study specifically testing how similar text generated by ChatGPT is to text produced by human writers.

\section{Methodology}

To investigate the trustworthiness of children's stories generated by LLMs, we compare them with actual old and modern children's stories.
We collect a diverse set of stories from different sources, including both older stories such as folktales, and more recent children's stories.
We use both LLMs and instruction-following models to generate stories with different prompt lengths and instruction templates.
As story generation is an open-ended problem with no reference text, we rely on other metrics instead of any automatic measure of evaluation like BARTScore~\cite{yuan2021bartscore} or BERTScore~\cite{zhang2019bertscore}.
We use various metrics to compare the generated stories with actual stories, including in-text statistics such as sentence length and a measure of toxicity in the text, as well as an evaluation of topics covered in these stories.
Furthermore, we analyze and compare the grammatical structures of the stories using dependency structures extracted from both the original and the artificially generated stories.

In the following section, we describe the experimental setup, including details on the collected data, the story generation process, and the evaluation metrics used for comparison.
Subsequently, we present the results obtained from our experimentation.
\section{Experiments}

\subsection{Data}

Our data consists of 132 original children's stories collected from various online sources and categorized into two categories: old and modern.
The old stories generally include traditional children's stories like folktales and fairy tales, whereas the modern stories include more recent children's literature published after the year 2000.
Both sets of original children's stories are comprised of English texts aimed at children between the ages of three and thirteen, with both data sets representing the full range of these target ages.
Overall, 122 are classified as old stories, and the remaining 10 as modern stories.
Specifically, the older stories were obtained via Project Gutenberg,\footnote{\href{https://www.gutenberg.org/}{https://www.gutenberg.org/}} and the modern stories from various online platforms.\footnote{\href{https://www.freechildrenstories.com/}{https://www.freechildrenstories.com/}, \href{https://monkeypen.com/}{https://monkeypen.com/}}
We use the old stories as a reference for the story generation task and compare the generated stories against both old and modern stories.

\subsection{Story generation}

We generate stories using language models and an instruction-following model.

\paragraph{Language Models}

Our story generation task using LLMs uses two foundational language models: OPT~\cite{zhang2022opt} and LLaMA~\cite{touvron2023llama}, with model sizes of 6.7 billion and 7 billion parameters, respectively.
To generate stories, we provide a portion of each old story as context for the LLMs.
Specifically, we use the first sentence, the first 256 tokens, and the first 512 tokens of each old story as a prompt.
We use top-$k$ sampling-based decoding with $k$ set to 100 and generate five samples for each prompt, resulting in a total of 3660 generated stories. The breakdown of the generated stories along with the length of the prompt is given in Table~\ref{tab:stat_lm}.

\begin{table}[t]
\centering
\small
\begin{tabular}{llr}
\toprule
Model & Prompt Length & Count\\
\midrule
OPT & First Sentence (OPT-Line) &  610\\
& First 256-tokens (OPT-256) &  610\\
& First 512-tokens (OPT-512) &  610\\
LLaMA & First Sentence (LLaMA-Line) &  610\\
& First 256-tokens (LLaMA-256) &  610\\
& First 512-tokens (LLaMA-512) &  610\\
\midrule
& \textbf{Total} &  3660\\
\bottomrule
\end{tabular}
\caption{Breakdown of the stories generated using LLMs.} 
\label{tab:stat_lm}
\end{table}

\paragraph{Instruction-following Models}

For instruction-following story generation, we use Alpaca~\cite{Alpaca}, which is an instruction-following model that is based on the LLaMA architecture and is fine-tuned using self-instruct~\cite{wang2022self}.
We use the Alpaca model based on the 7B variant of the LLaMA model.
We use four different instruction templates to generate stories, two of which require a story title as input and two of which do not. 
For the templates that require a story title, we use the title of old stories as input.
The templates are provided in Table~\ref{tab:templates}.
To generate stories, we use top-$k$ sampling-based decoding with $k$ set to 100 and generate five samples for each template, resulting in a total of 2440 generated stories with 610 stories per template.

\begin{table*}[t]
\centering
\small
\begin{tabular}{ll}
\toprule
S.N. & Template\\
\midrule
T1 &  Below is an instruction that describes a task, paired with an input that provides further context. \\
& Write a response that appropriately completes the request. \\
& \\
& \#\#\# Instruction: \\
& Write a short children's story given the title. \\
& \\
& \#\#\# Input: \\
& {TITLE} \\
& \\
& \#\#\# Response: \\
\midrule
T2 &  Below is an instruction that describes a task. Write a response that appropriately completes the request.\\
& \\
& \#\#\# Instruction: \\
& Write a short children's story. \\
& \\
& \#\#\# Response: \\
\midrule
T3 &  Below is an instruction that describes a task, paired with an input that provides further context. \\
& Write a response that appropriately completes the request. \\
& \\
& \#\#\# Instruction: \\
& Write a children's story given the title. \\
& \\
& \#\#\# Input: \\
& {TITLE} \\
& \\
& \#\#\# Response: \\
\midrule
T4 &  Below is an instruction that describes a task. Write a response that appropriately completes the request.\\
& \\
& \#\#\# Instruction: \\
& Write a children's story. \\
& \\
& \#\#\# Response: \\

\bottomrule
\end{tabular}
\caption{Templates used by Alpaca for story generation.} 
\label{tab:templates}
\end{table*}

\subsection{In-text statistics}
We compare various statistics derived from the text of the generated stories against those of actual stories.
Specifically, we use two metrics: sentence length and Flesch reading ease score~\cite{flesch1948new}.

\paragraph{Flesch Reading Ease Score } 
The Flesch reading ease score (FRES) measures the readability of a text and is based on two factors: average sentence length and the average number of syllables per word.
It provides a score between 0 and 100, with higher scores indicating easier readability.
A Flesch reading ease score above 60 for a text indicates that it can easily be read by children up to the age of 15.
The formula for calculating the FRES of a text is shown in Equation~\ref{eq_fres}.

\begin{multline}
\label{eq_fres}
    FRES = 206.835 - 1.015 \left( \frac{total \; words}{total \; sentence} \right)  \\
    - 84.6 \left( \frac{total \; syllables}{total \; words} \right)
\end{multline}

\subsection{Toxicity of text}
\citealt{gehman-etal-2020-realtoxicityprompts} found that the LLMs can generate `toxic' text from a very innocuous prompt and attribute this to a significant amount of offensive, factually unreliable, and otherwise toxic content in the training data of these models.
We want to investigate the level of toxicity in our generated children's stories.
Ideally, generated children's stories should be free of any toxic text.

We use Detoxify~\cite{Detoxify}, a BERT~\cite{devlin-etal-2019-bert} based toxic text detector, to identify the presence of toxic text in the generated children's stories.
Detoxify generates score labels in the range of 0 to 1, assessing the toxicity of the text based on categories such as toxic, severely toxic, obscene, threat, insult, and identity hate.
Specifically, we use detoxify for each sentence of our actual and generated stories to get toxicity measures across the six categories.

\subsection{Topic Modeling}

We also analyze the data for topic modeling using pyLDAvis~\cite{tran_pyldavis_2022}.
We compare the topics found in the data set of older stories with the LLM-generated stories.
The older stories and the modern stories are also compared to assess whether there has been a shift in topics over time that would potentially influence topic properties in the LLM-generated stories.
A probable diachronic shift in topics of stories geared towards young audiences also highlights the need to test the toxicity of generated stories, as seen in the previous section.

To avoid uninformative topics, the data is preprocessed to remove stopwords and names.
All texts are categorized for specific topics using word clustering for a set of documents.
Modeling is performed automatically without a predefined list of labels.
The visualizations using pyLDAvis break down the topics based on the 122 older stories, the 10 modern stories, and the generated stories from OPT, LLaMA, and Alpaca.

\subsection{Sentence structure}

The structure of the sentences within a text can reveal the type or genre of the text.
To analyze sentence structures, we construct a dependency tree for each sentence in both the original and generated children's stories.
The dependency tree depicts the syntactic dependencies between the words in a sentence, effectively capturing the grammatical structure of the sentence.
We then convert these dependencies into unlabeled directed graphs, preserving sentence structure while removing specific words.
We then generate the Weisfeiler Lehman graph hash~\cite{shervashidze2011weisfeiler} for each graph. 
The Weisfeiler Lehman hashes are identical for isomorphic graphs and strongly guarantee that non-isomorphic graphs will get different hashes. 
We compare the frequency of hashes to evaluate the similarity between the sentence structure of the generated stories and the actual stories.

\section{Generated stories follow modern trends but struggle with nuances}

\begin{figure}
    \centering
    \includegraphics[width=\linewidth]{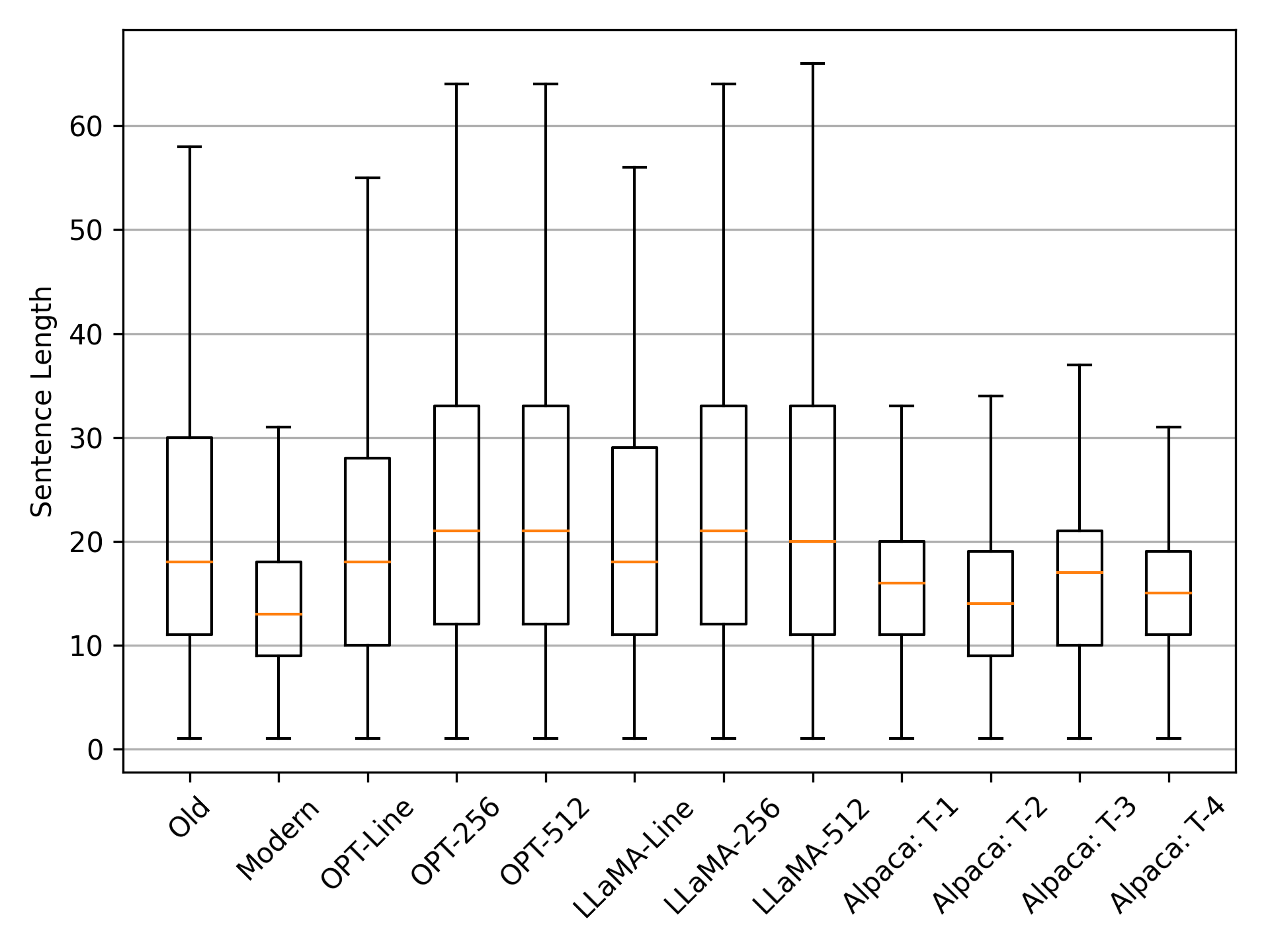}
    \caption{Comparison of sentence length in generated children's stories and actual children's stories. The generated children's stories exhibit shorter sentence lengths compared to the older original stories but are similar in sentence length to modern stories. Language models prompted with older stories tend to generate longer sentences following the patterns of the context that had been provided.}
    \label{fig:sen_lengths}
    \vspace{-1em}
\end{figure}

Figure~\ref{fig:sen_lengths} shows the box plot of sentence lengths for old and modern original stories, as well as for the generated stories.
Being literary texts, children's stories do not strictly confine to formal English conventions and many contain sentences with higher word counts; so for clarity, we removed all the outliers from the plot.
One interesting observation is that modern children's stories generally have shorter sentence lengths than older children's stories, which adheres to  previous research that shows a trend of decreasing sentence length in print~\cite{ctx20685388910004105}.
The generated stories from OPT and LLaMA show an increase in sentence length as the prompt length increases. 
We hypothesize that these models learn the pattern of larger sentence length from the older stories used as context, which is then reflected in the generated text.
However, stories generated using the instruction-following model Alpaca, have sentence lengths similar to modern actual stories, indicating that language models may have been trained mostly on the newer text, and tend to generalize modern trends when instructed to generate text of a specific type.

\begin{figure}[t]
    \centering
     \begin{subfigure}[b]{0.45\textwidth}
         \centering
         \includegraphics[width=\textwidth]{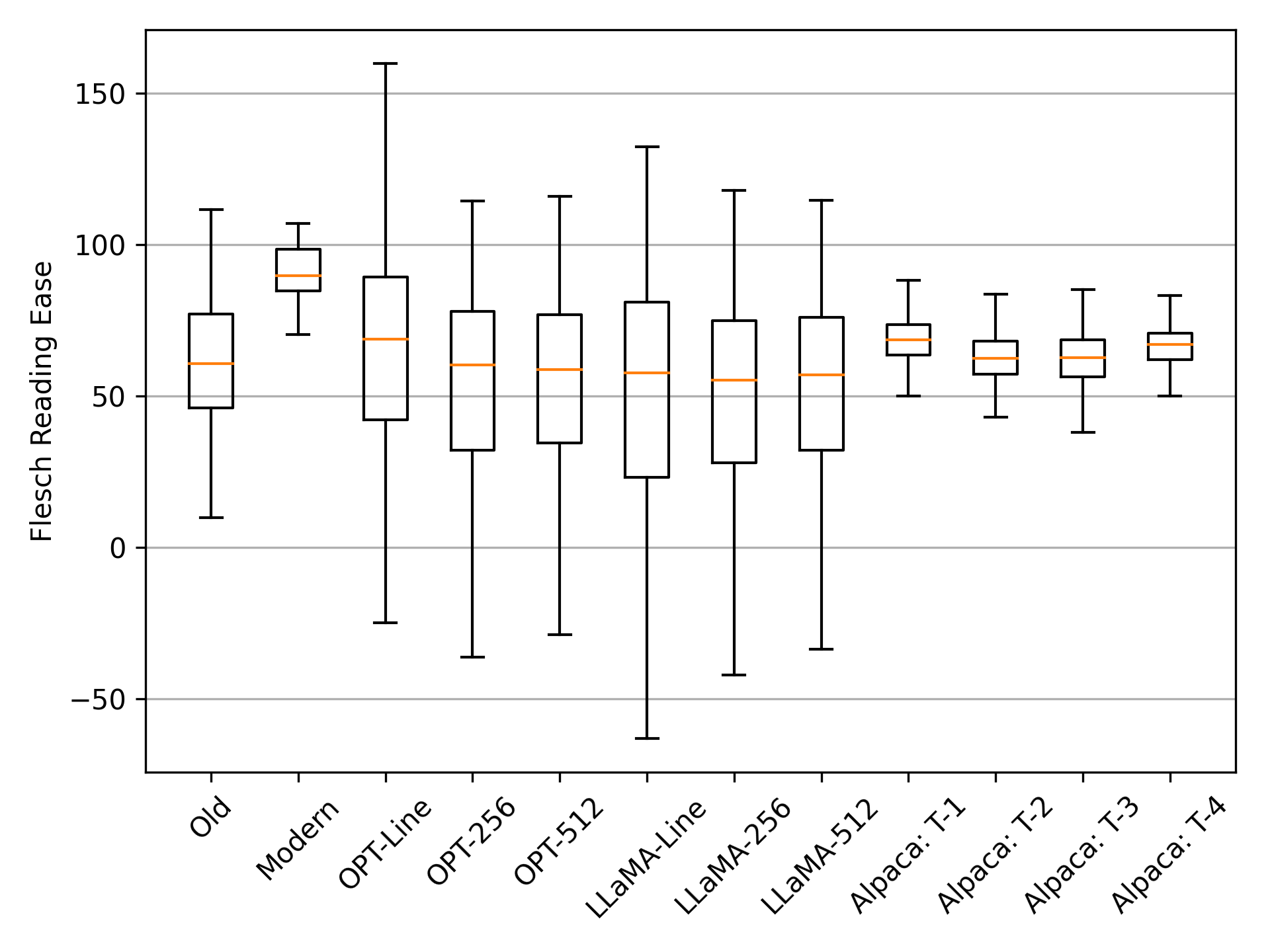}
         \caption{FRES on all data (over 100 is undefined).}
         \label{fig:readibility_score_}
     \end{subfigure}
     \begin{subfigure}[b]{0.45\textwidth}
         \centering
         \includegraphics[width=\textwidth]{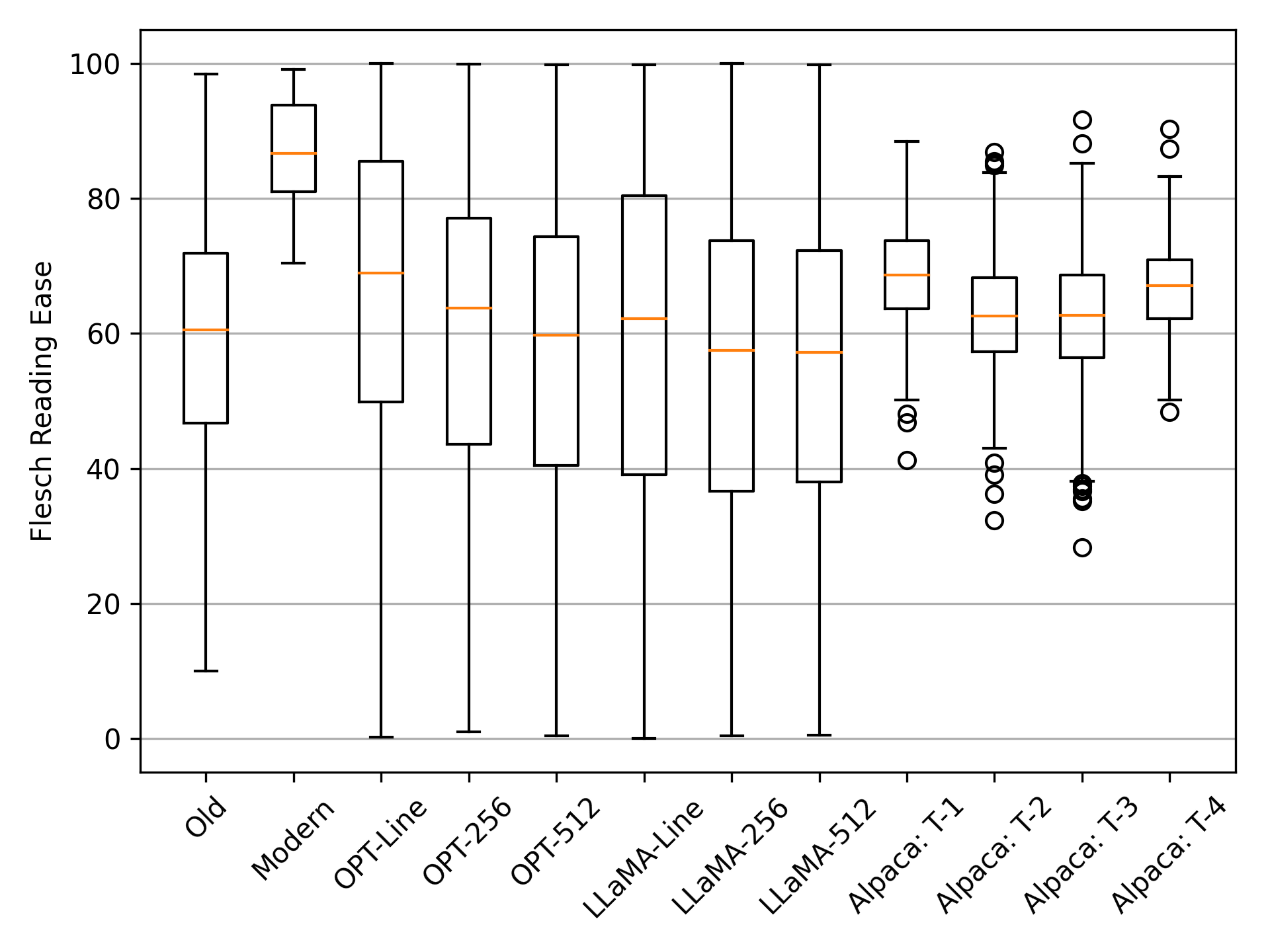}
         \caption{FRES limited to well-defined 0-100 range.}
         \label{fig:readibility_score_range}
     \end{subfigure}
    \caption{Comparison of FRES in generated children's stories and actual children's stories : (a) FRES with all data and (b) FRES only in the range of 0 and 100. 
    The generated children's stories are easier to read compared to older actual stories but are not as easy as modern original stories. Language models prompted with older stories tend to generate text that is more difficult to read, likely because they follow the patterns in the prompts. }
    \label{fig:readibility_score}
    \vspace{-1em}
\end{figure}

The Flesch reading ease score is a statistical measure of the readability of a text and was optimized to be general enough at the time of its formulation, as can be seen with the constant values in equation~\ref{eq_fres}.
That is why, we may find FRES values not within the range of 0 to 100 as seen in Figure~\ref{fig:readibility_score_}.
We also removed the outliers from the box plot in Figure~\ref{fig:readibility_score_}.
Since we are not interested in exact values but in the general trend these values represent, we use the FRES values in the range of 0-100 and show their box plot in Figure~\ref{fig:readibility_score_range}.

Our results from the Flesch reading ease score reveal several interesting observations.
Firstly, we see that modern children's stories have a higher FRES than older stories, meaning that the modern ones are easier to read.
This can be attributed to the fact that sentences are getting shorter and might have to do with simpler word selection.
Secondly, we see that LLMs prompted with older stories tend to follow the pattern of the context and generate stories that are more difficult to read, as the context length increases.
Finally, we see that the instruction-following model Alpaca generates stories that are easier to read compared to older original children's stories but are not as readable as modern children's stories.
We posit that this observation can be attributed to the fact that LLMs used in our study are generic models, and the instruction following model is also only fine-tuned for general instructions rather than instructions specific to children's story generation.

\begin{figure*}[t]
    \centering
     \begin{subfigure}[b]{0.32\textwidth}
         \centering
         \includegraphics[width=\textwidth]{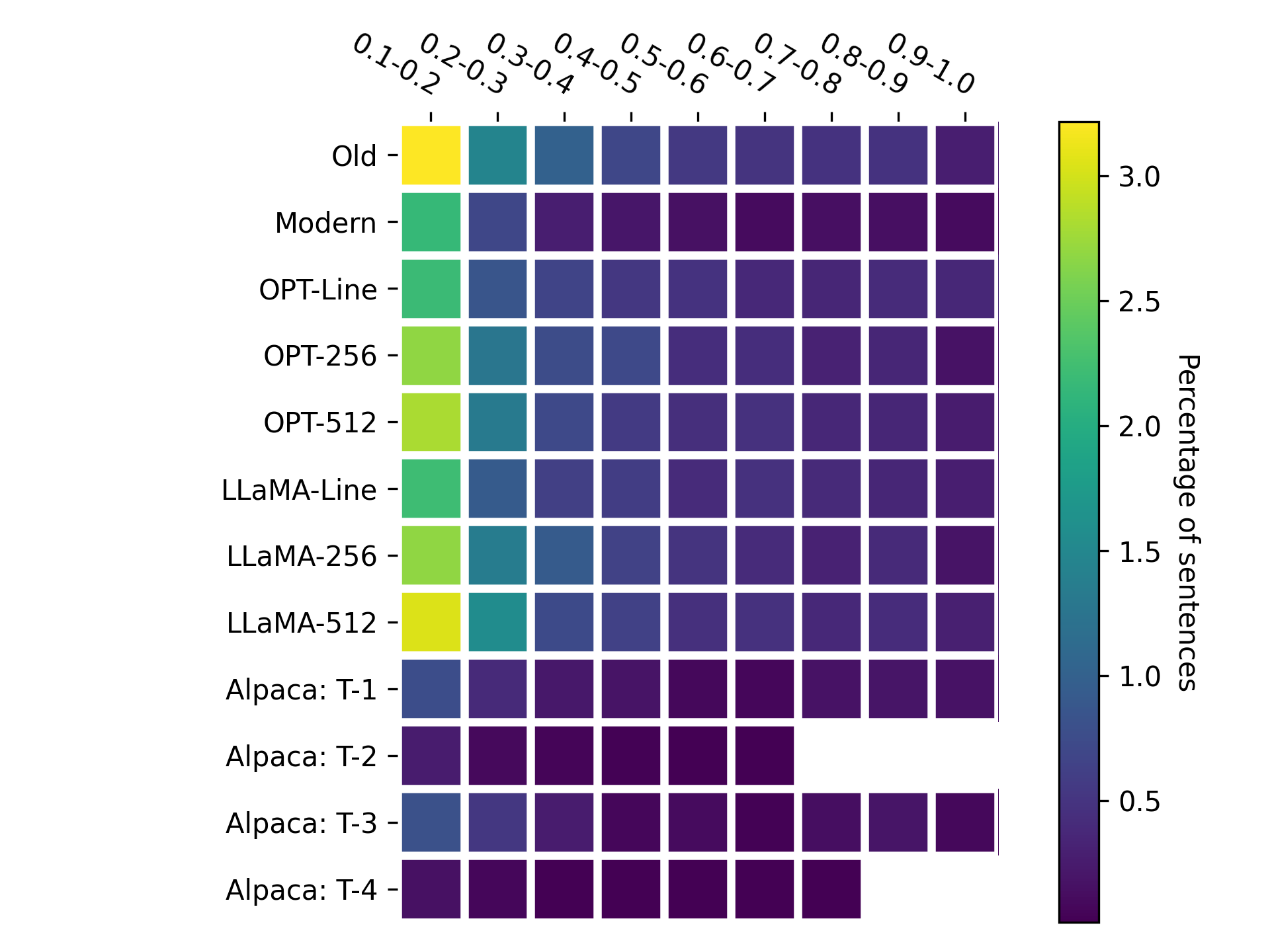}
         \caption{toxic}
         \label{fig:toxic_measures_toxic}
     \end{subfigure}
     \begin{subfigure}[b]{0.32\textwidth}
         \centering
         \includegraphics[width=\textwidth]{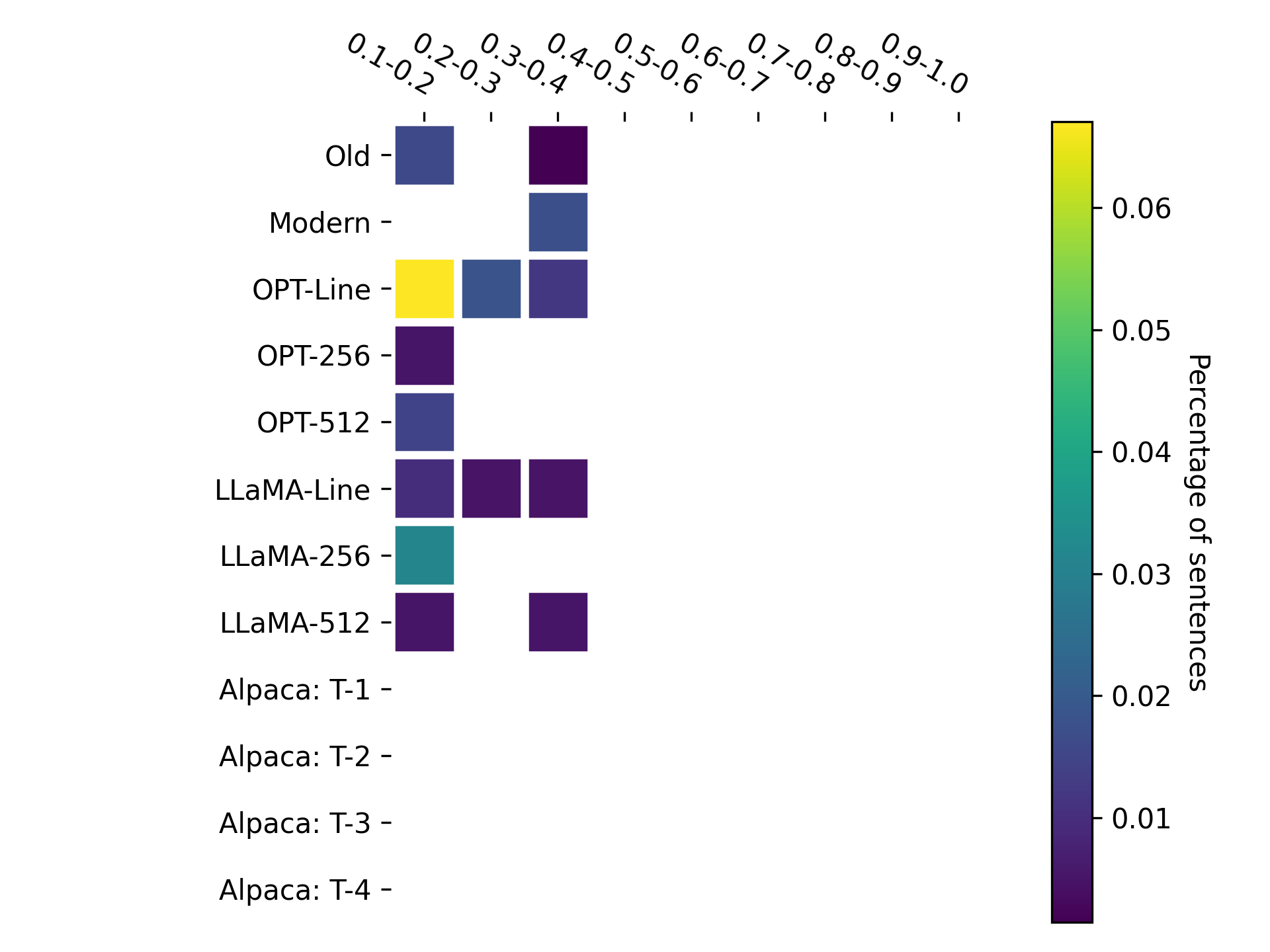}
         \caption{severe toxic}
         \label{fig:toxic_measures_severe}
     \end{subfigure}
    \begin{subfigure}[b]{0.32\textwidth}
         \centering
         \includegraphics[width=\textwidth]{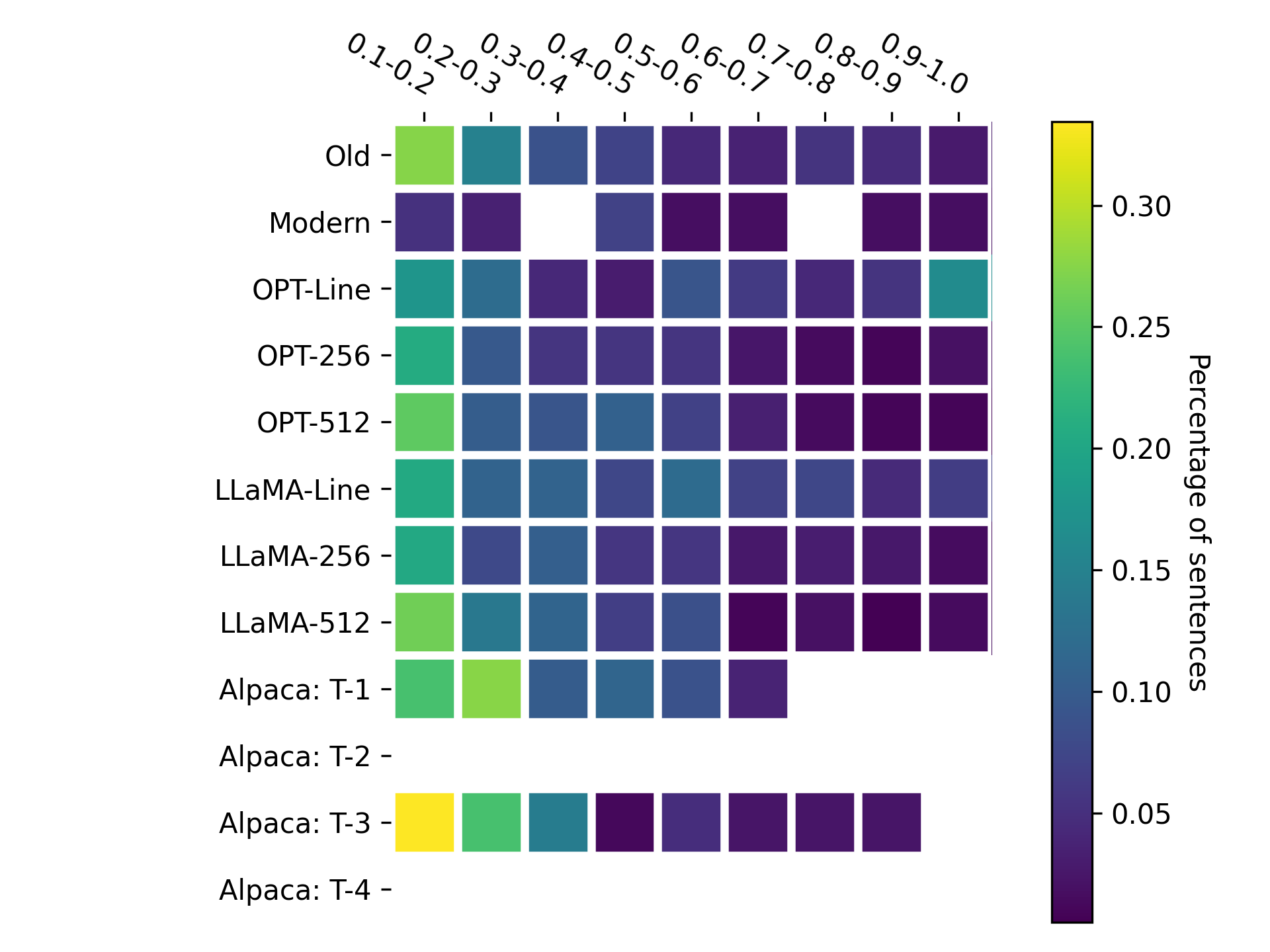}
         \caption{obscene}
         \label{fig:toxic_measures_obscene}
     \end{subfigure}
     \begin{subfigure}[b]{0.32\textwidth}
         \centering
         \includegraphics[width=\textwidth]{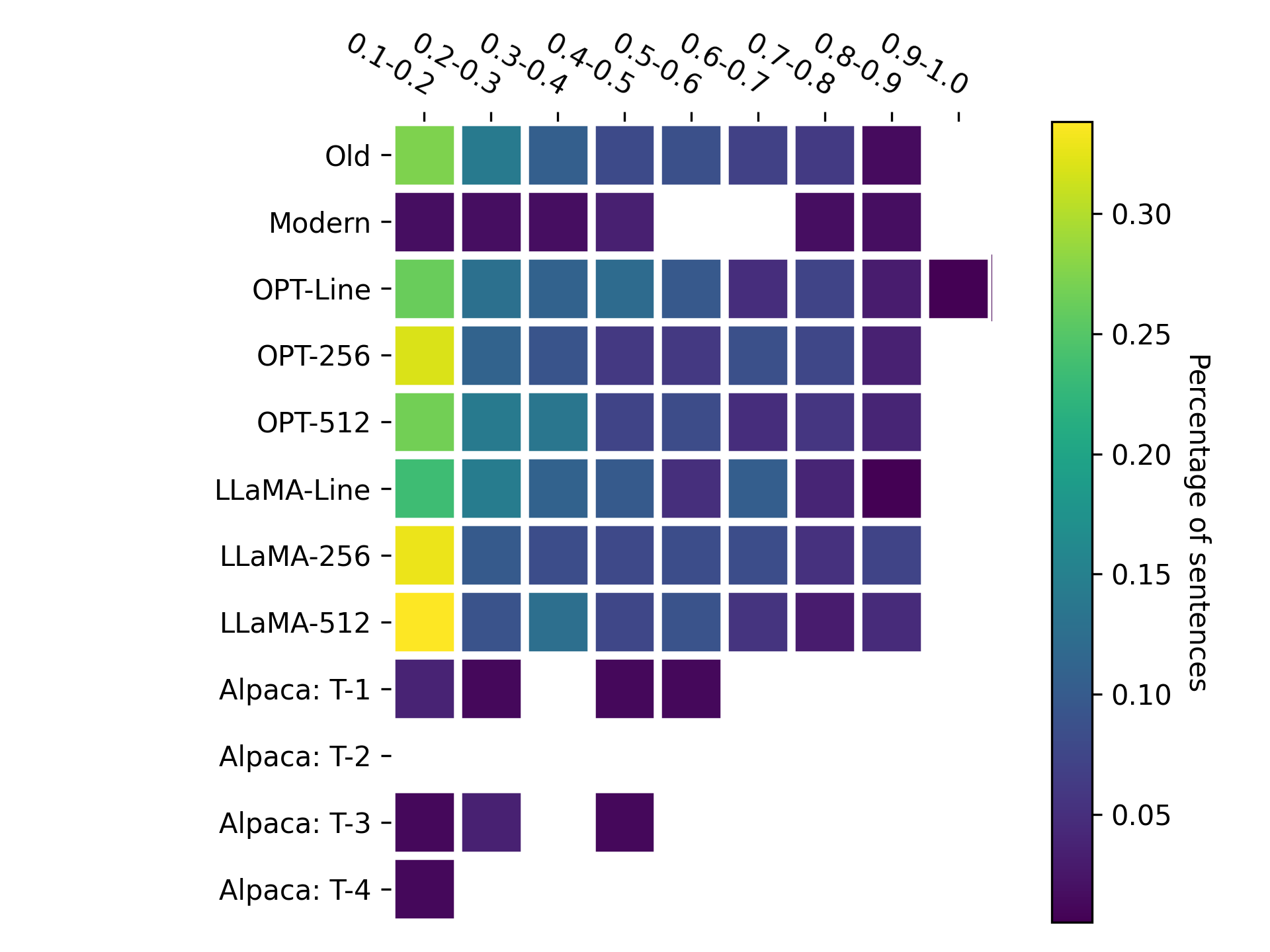}
         \caption{threat}
         \label{fig:toxic_measures_threat}
     \end{subfigure}
     \begin{subfigure}[b]{0.32\textwidth}
         \centering
         \includegraphics[width=\textwidth]{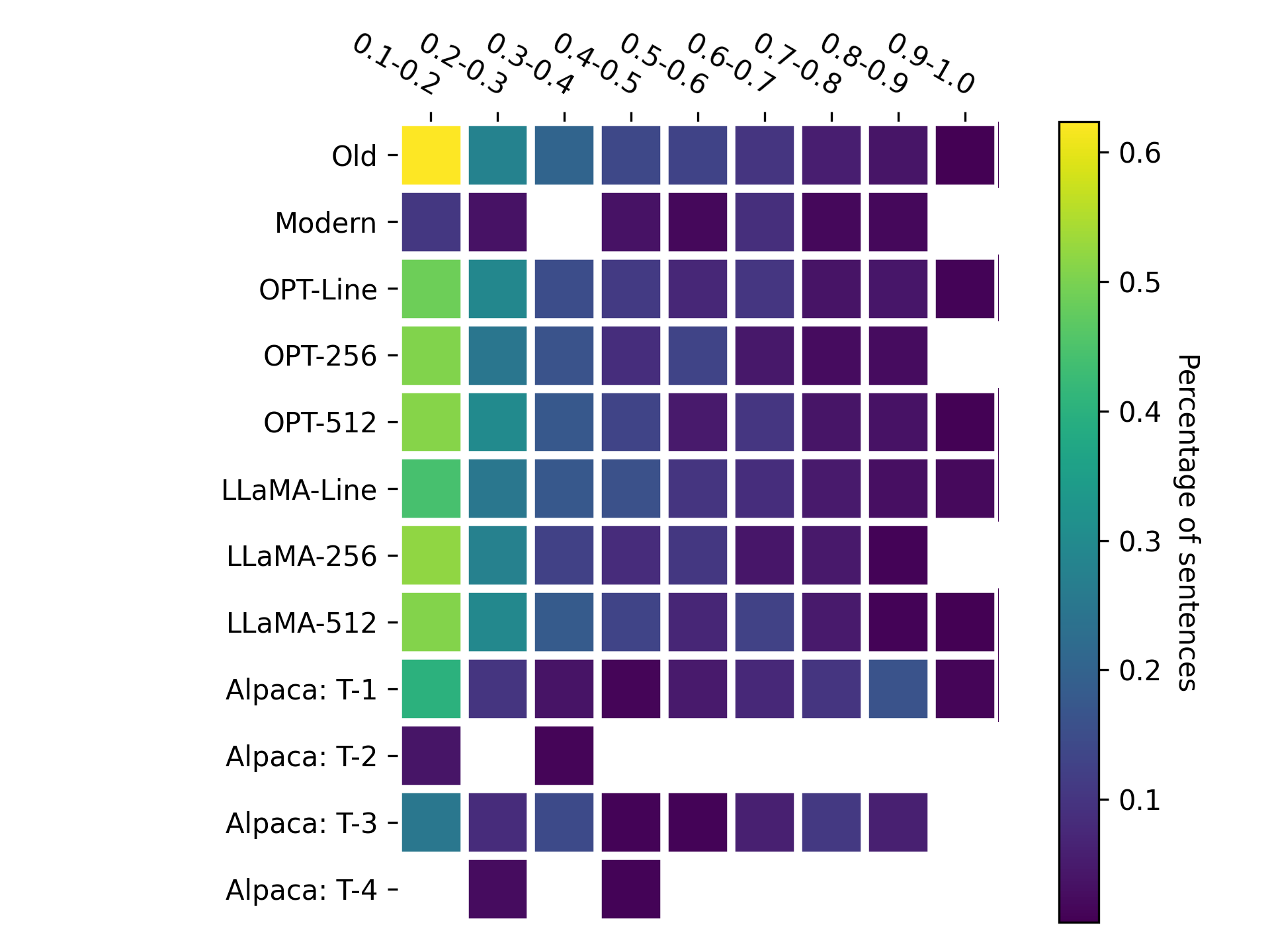}
         \caption{insult}
         \label{fig:toxic_measures_insult}
     \end{subfigure}
    \begin{subfigure}[b]{0.32\textwidth}
         \centering
         \includegraphics[width=\textwidth]{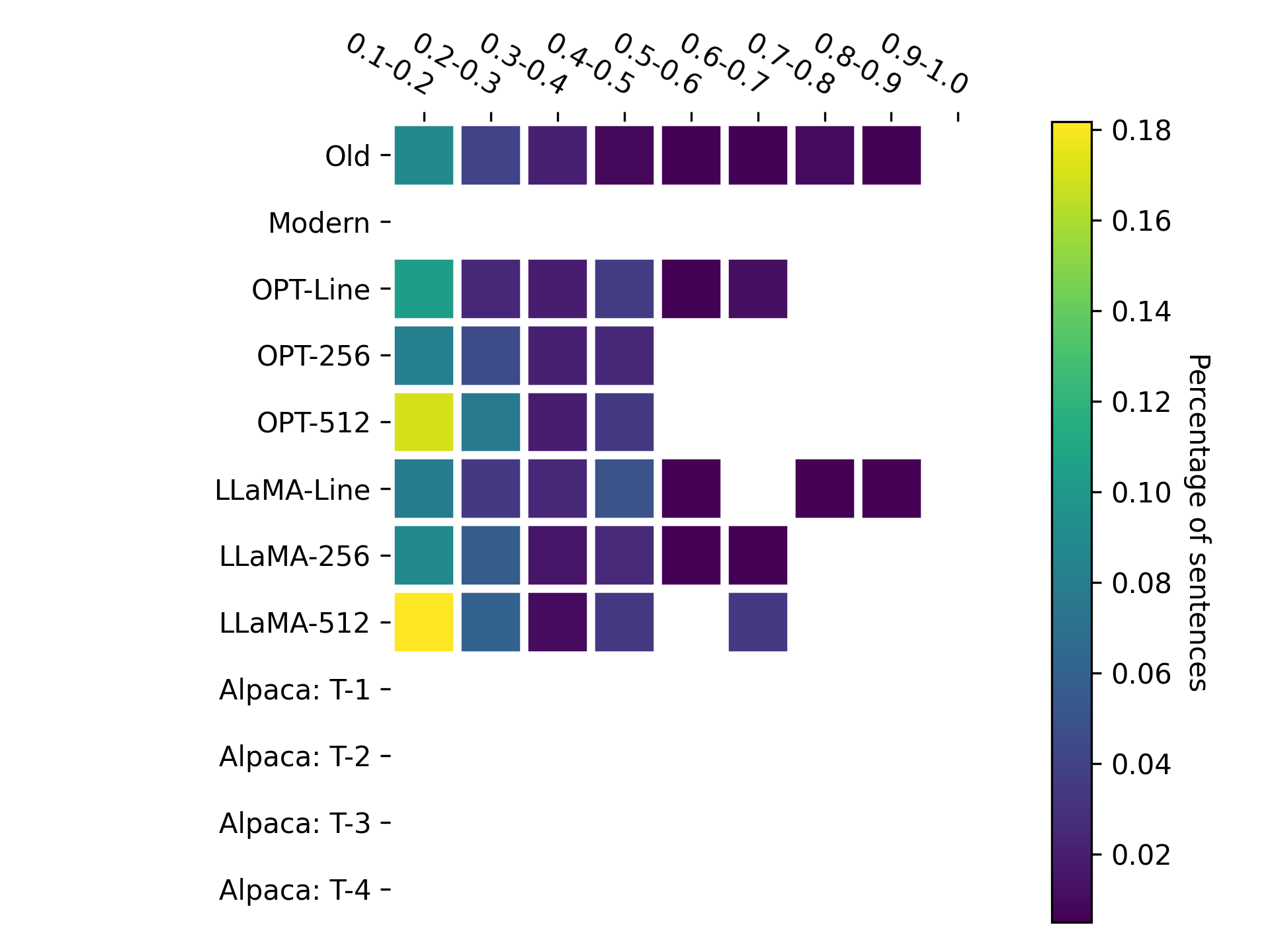}
         \caption{identity hate}
         \label{fig:toxic_measures_identity}
     \end{subfigure}
    \caption{Various toxicity measures for the actual and generated stories. Each cell in a subplot represents the percentage of sentences rated on a toxicity scale, with x-axis values indicating the toxicity level. Values for ratings in the range of 0-0.1 have been omitted from the plots for clarity.}
    \label{fig:toxic_measures}
\end{figure*}

Overall, we see that modern children's stories are easier to read than older children's stories. 
As most of the training data for LLMs comes from newer text, the model tends to follow the trend of modern children's stories in their generated text for sentence length and word selection. 
However, it should be noted that these models are not fine-tuned for children's stories generation, and therefore may not capture the nuances of children's stories resulting in stories that might be difficult to read for intended readers.

\section{Generated stories may contain toxic text}
Our analysis of toxicity in actual and generated stories reveals several noteworthy findings.
We present the toxicity measures for both actual and generated stories in Figure~\ref{fig:toxic_measures}. 
Notably, we find that older stories tend to be more toxic than modern stories across all toxicity measures.
This trend is not solely due to the smaller sample size of modern actual stories, as we have normalized the toxicity ratings to ensure an accurate comparison.
Rather, it suggests that writers are becoming more mindful of the language they use in children's literature.
Although modern stories are less toxic compared to older stories, we still observe some level of toxicity in them.
This toxicity in modern actual stories is often related to the narrative of the story.
For example, threats and insults might be needed for some stories, but identity hate is not appropriate for children's stories.
It is noteworthy that modern stories do not have toxic text related to identity hate but older stories do.

Similar to our previous observation, we see that LLMs tend to learn patterns from the context they are provided with.
As evident from the stories generated by OPT and LLaMA, we see that the toxicity aligns with older stories and gradually increases with an increase in the length of the context.
The stories generated using the instruction-following model Alpaca tend to be less toxic and mostly resemble modern stories.
However, stories generated using the T1 and T3 templates have a lot of obscene text compared to stories generated using T2 and T4, which have none.
As shown in Table~\ref{tab:templates}, T1 and T3 take the title as input whereas T2 and T4 do not.
It is possible that the model remembered the story title and generalized the patterns of the story or generalized to some other text in the template, leading to the generation of obscene text.
This finding is consistent with \citealt{gehman-etal-2020-realtoxicityprompts}, who suggest that children's stories generated by LLMs can contain highly toxic text despite an innocuous prompt.

Our analysis of toxicity in original and generated stories reveals that older stories tend to be more toxic than modern ones, that LLMs can learn toxic patterns from context leading to the generation of toxic text, and that LLMs can even generate toxic text from a very innocuous prompt. These findings suggest that further work is needed to make LLMs useful as tools for generating age-appropriate children's literature.

\section{Generated stories share main topics with original stories}

After preprocessing the data, the original stories were found to have four major topics.
All of the topics tended to share the existence of some small character.
The first topic mentions elements such as time, goodness, and greatness, and the presence of words like head, round, night, and water likely indicate specific scenes or settings within the narrative.
The second topic contained new elements like a prince, the color white, a girl, and eyes.
These additional keywords suggest different perspectives within the overarching narrative.
The third topic introduces elements like a house and a heart.
Like the previous topics, it shares mentions of a little character, time, goodness, and a prince.
The difference between 'house' and 'heart' could indicate a change in the setting or moral of the narrative.
The last topic introduces new elements of wolf, people, eyes, and a mother.
These keywords might suggest narratives that introduce new characters and themes.
Overall, these topics provide insight into the underlying themes present in the older 122 stories in the data set.
The topics revolve around narratives involving a small character, time, goodness, and various other elements such as princes, nights, water, girls, and wolves.

Comparatively, the topics of the generated stories obtained from OPT, LLaMA, and Alpaca show minor differences. 
The first topic suggests a narrative that involves characters like kings, mothers, princes, and princesses.
It also mentions elements of time, goodness, greatness, and shadow.
The prince, princess, and shadow hint at the fairy tale or fantasy theme.
The second topic shares similarities with the previous topic, with a focus on little, prince, time, goodness, and greatness, but it also introduces new elements like eyes, houses, heads, and the color white.
These additions suggest different scenes, perhaps removed from the monarchy or castle theme, and suggest a different narrative.
The third topic seems to center around family dynamics, with mentions of mothers, fathers, and children. It also includes keywords relating to time, goodness, night, and poverty. This suggests a change in the narrative away from the fantasy-focused topic.
The last topic includes keywords like little, time, and goodness. It includes elements of fathers, eyes, and houses. The presence of 'long' and 'night' suggests a different tone or atmosphere within the narrative.
These general results show remarkable similarity with the data set on which the LLMs were trained.
The topics revolve around narratives involving characters such as kings, princes, mothers, fathers, and children.
The topics also touched upon topics of time, goodness, greatness, poverty, and setting elements of houses, nights, and the color white.

\begin{figure}
    \centering
    \includegraphics[width=\linewidth]{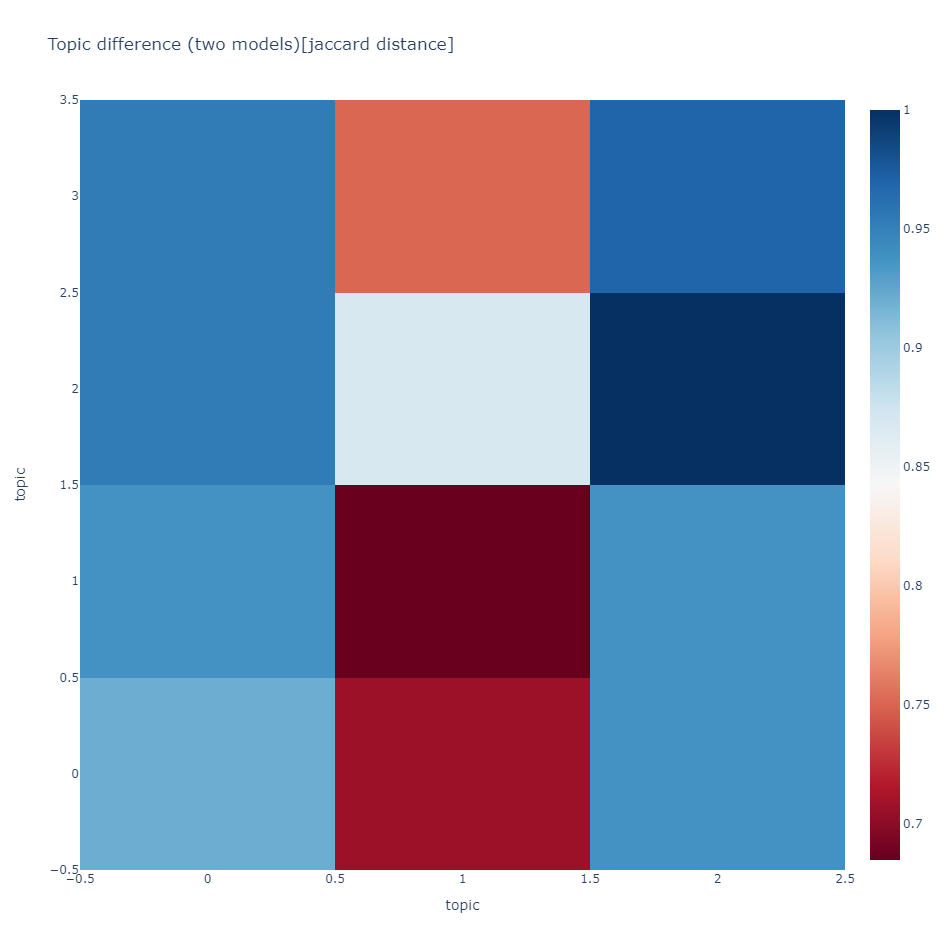}
    \caption{Comparison of topics in generated children's stories and actual children's stories. The plot shows that the most shared topics (x:2, y:2) include 'white', 'world', 'great', 'water', 'black', 'house', 'little', 'king', 'called', and 'good'. The least shared topics (x:1, y:1) include 'heart', 'head', 'poor', 'house', 'looking', 'children', 'good', 'young', 'lady', and 'night'.}
    \label{fig:topic_modeling}
\end{figure}

As with toxic content testing, we ran topic modeling for a small number (10) of modern stories in order to compare the general topics that are currently aimed at children.
The first topic includes keywords related to spatial orientation (right, inside, door, left), objects (ream, head, frog), time, and actions (started).
The keyword `eyes' may suggest a focus on visual perception or observation.
The second topic emphasizes time, objects (ream, door), spacial orientation (right, inside), a frog, a head, fairies, and `need'.
The presence of fairies introduces a fantastical or imaginative element to the topic.
The third topic revolves around time, spacial orientation (right, inside, door), physical attributes (head, eyes, long, hand), and a frog.
The inclusion of `long' might suggest a temporal or duration-related aspect.
The last topic highlights time, spatial orientation (right, inside, door), objects (ream, frog), physical attributes (head, eyes, small), and the action of starting something.
The modern stories' topic modeling results suggest a recurring theme involving concepts such as time, spatial orientation, objects, and actions.
Each topic emphasizes different aspects and introduces additional elements like fairies or physical attributes. Figure~\ref{fig:topic_modeling} represents the level of similarity and difference between the real stories and the LLM-generated stories.
There are greater similarities between these stories than there appear to be differences in the main topics.

As expected, the results of the topic modeling showed similarities between the original 122 stories in the training corpus and the stories generated by the LLMs.
These stories shared fairy tale and fantasy elements as well as topics of goodness, greatness, time, and setting elements of night, houses, and the color white.
Once we compare this with the modern stories, we see that the focus of the small data set we have is similarly focused on time and fairies, but has more topics relating to spatial orientation.
We are likely seeing a change in the content of stories written for children.
With only ten modern stories, we cannot reliably generalize over all stories, but we noticed tendencies such as that the modern story set did tend to involve more overtly educational elements aimed at younger age groups when compared to the older stories.

\section{Generated stories do not have similar sentence structure to original stories}
Table~\ref{tab:dependency_tree} shows the percentage of overlapping Weisfeiler Lehman hashes between the dependency tree graphs of sentences generated by various models and those actual children's stories, both old and modern.
We also got an overlap of \textbf{35.57} percentage between old and modern actual stories, which is greater than all the values in Table~\ref{tab:dependency_tree}.
This shows that the structure of sentences in children's literature has changed over time, which supports our earlier findings that children's literature has undergone noticeable changes over time.
\begin{table}[h]
\centering
\begin{tabular}{lrr}
\toprule
\multirow{2}{*}{Model} & \multicolumn{2}{c} {Percentage overlap with} \\
 &  Old stories & Modern stories \\
\midrule
OPT-Line & 34.82 & 34.21\\
OPT-256 & 31.37 & 28.88\\
OPT-512 & 32.49 & 29.89\\
LLaMA-Line & 34.23 & 33.64\\
LLaMA-256 & 32.14 & 29.82\\
LLaMA-512 & 32.27 & 30.73\\
Alpaca: T-1 & 17.31 & 20.37\\
Alpaca: T-2 & 14.67 & 17.52\\
Alpaca: T-3 & 15.20 & 16.92\\
Alpaca: T-4 & 15.41 & 17.84\\
\bottomrule
\end{tabular}
\caption{Overlap of the hashes of the dependency tree graph of the sentences in generated stories against old and modern actual stories.} 
\label{tab:dependency_tree}
\end{table}

Additionally, we observe a higher percentage of overlap between old original stories and the stories generated by OPT and LLaMA, which again aligns with our earlier findings that LLMs learn from their context.
Furthermore, for the stories generated by OPT and LLaMA, we see an average overlap of 30\% with modern stories, which can be attributed to the fact that these models were trained on a dataset consisting of recent text.

The stories generated by Alpaca have a slightly higher overlap with modern stories compared to old stories, but the percentage overlap in sentence structures is still relatively low ( \(\le\) 20\%).
Given that the old and modern actual stories share around 35\% of the same sentence structures, we expected Alpaca's generated stories to overlap more with modern stories.
But since Alpaca is a generic model fine-tuned for instruction-following and not solely trained or fine-tuned on children's literature, it seems plausible that it would not be capable of fully generalizing over sentence or grammatical structures observable in children's literature.

\section{Conclusion and Future Work}
Our study examines the trustworthiness of children's stories generated by large language models. 
While these generated stories may share similar topics and patterns with actual stories, they fail to capture all the nuances present in children's literature, and may even contain toxic material that is inappropriate for children.
Based on our findings, we conclude that LLMs are not yet appropriate for generating high-quality children's literature.
Moving forward, we plan to extend our work by implementing reinforcement learning with both automatic and human feedback to improve the quality of LLM-generated children's stories.

\section*{Acknowledgments}

This work was supported by resources provided by the Office of Research Computing, George Mason University (URL: https://orc.gmu.edu) and by the National Science Foundation (Awards Number 1625039 and 2018631).
Additionally, Prabin Bhandari has been partially supported by the National Science Foundation Grant No. IIS-2127901.
Our sincere appreciation goes to Antonios Anastasopoulos and Géraldine Walther for their valuable suggestions and unwavering support throughout the course of this research.

\bibliographystyle{acl_natbib}
\bibliography{anthology,acl2021}

\appendix

\end{document}